\documentclass[conference]{IEEEtran}
\IEEEoverridecommandlockouts
\usepackage{cite}
\usepackage{amsmath,amssymb,amsfonts}
\usepackage{algorithmic}
\usepackage{graphicx}
\usepackage{textcomp}
\usepackage{algorithm}
\usepackage{algorithmic}
\usepackage{xcolor}
\def\BibTeX{{\rm B\kern-.05em{\sc i\kern-.025em b}\kern-.08em
    T\kern-.1667em\lower.7ex\hbox{E}\kern-.125emX}}
\begin{document}

\title{Spatio-Temporal Pruning and Quantization for Low-latency Spiking Neural Networks\\
}

\author{\IEEEauthorblockN{Sayeed Shafayet Chowdhury, Isha Garg and Kaushik Roy}
\IEEEauthorblockA{\textit{School of Electrical and Computer Engineering} \\
\textit{Purdue University}\\
West Lafayette, IN 47905, USA \\
\{chowdh23, gargi, kaushik\}@purdue.edu}

}

\maketitle

\begin{abstract}

Spiking Neural Networks (SNNs) are a promising computationally  efficient alternative to traditional deep learning methods since they perform sparse event-driven information processing. 
However, a major drawback of using SNNs is the high inference latency, specially for real-time applications. The efficiency of SNNs could be further enhanced using compression methods such as pruning and quantization. Notably, SNNs, unlike their non-spiking counterparts, consist of a temporal dimension, the compression of which can lead to significant latency reduction. In this paper, we propose spatial and temporal pruning and quantization of SNNs. First, structured spatial pruning is performed by determining the layer-wise significant dimensions using principal component analysis of the average accumulated membrane potential of the neurons. This step leads to 10-14X model size compression. Additionally, the spatially-pruned SNNs enable inference with lower latency and decrease the total spike count per inference. To further reduce latency, temporal pruning is performed by gradually reducing the timesteps of
simulation while training. The networks are trained using surrogate gradient descent based backpropagation and we validate the results on CIFAR10 and CIFAR100, using VGG architectures. The proposed spatio-temporally pruned SNNs achieve 89.04\% and 66.4\% accuracy on CIFAR10 and CIFAR100, respectively, while performing inference with 3-30X reduced latency compared to other state-of-the-art SNNs. 
Moreover, they require 8-14X lesser compute energy compared to their unpruned standard deep learning counterparts. The energy numbers are obtained by multiplying the number of operations with energy per operation. These SNNs also provide 1-4\% higher accuracy in presence of Gaussian noise corrupted inputs. Furthermore, we perform post-training weight quantization and find the network performance remains reasonably stable up to 5-bit quantization.

\end{abstract}


\section{Introduction}
Over the past decade, the deep learning revolution has achieved impressive performance on many challenging tasks across domains such as computer vision and natural language processing \cite{b1}--\cite{b2}. However, such deep learning approaches are highly compute-intensive \cite{li2016evaluating} and therefore it remains a challenge to deploy them for resource-constrained applications. To tackle this, one approach is to use various model compression
techniques such as low-rank decomposition \cite{lowrank}, pruning \cite{han2015}--\cite{wen2016}, and data quantization \cite{binarynet}, which have provided a considerable boost in hardware performance \cite{haq}. Another route for efficient computation is employing Spiking Neural Networks (SNNs) \cite{mass}, which                      perform computation using spikes instead of the analog values used in traditional non-spiking deep neural networks, referred to as Analog Neural Networks (ANNs) henceforth. The energy efficiency of SNNs primarily stems from the sparsity of computation at any given timestep and from the replacement of multiply–accumulate (MAC) operations in the ANNs by additions. However, a key distinction between ANN and SNN is the notion of time; SNNs require computation over multiple timesteps, whereas ANNs infer in a single shot in the temporal dimension. If the latency requirements are too high for SNNs, it also impacts energy-efficiency negatively. 
As such, there is scope to further enhance the energy-benefits of SNNs if pruning and quantization can be incorporated suitably in them.

Several works have explored the applicability of compression methods for SNNs. Models with ternary weights and
binary activations were deployed on TrueNorth \cite{esser}. Dora \textit{et al.}~\cite{dora} proposed a two-stage growing-pruning method for fully-connected (FC) SNNs.  A
spike-timing dependent plasticity (STDP) learning rule with soft-pruning
achieved 95.04\% accuracy on MNIST 
\cite{shi}. Rathi \textit{et al.}~\cite{rathi} used pruning and weight quantization on FC SNNs with STDP, and obtained 91.5\% accuracy on MNIST. Yousefzadeh \textit{et al.~}\cite{yousefzadeh} reported 95.7\% accuracy
on MNIST using SNNs trained with 
stochastic STDP and reduced bit precision. Notably, these approaches were mostly limited to small-scale MNIST dataset. Srinivasan \textit{et al.~}\cite{gopal} introduced residual paths
into spiking convolutional (conv) layers
with binary weight kernels trained with probabilistic STDP  and demonstrated 98.54\% accuracy
on MNIST but only 66.23\% on CIFAR10. Additionally, the FC layers in this work were non-spiking. Recently, a comprehensive SNN compression approach was presented in \cite{admm} using the ADMM optimization tool, but most of the analysis still only focused on MNIST-like datasets. An attention-guided compression approach to limit spiking activity was proposed in \cite{spike-thrift}, but it cannot reduce the latency considerably, also the accuracy for CIFAR100 is lower in this method compared to other works.  Moreover, the feasibility of performing compression along the temporal axis of SNNs has not been explored in any of these prior works. Reducing timesteps holds great promise since such time-axis compression is directly correlated with latency as well as number~of spikes or energy consumption.

In this paper, we propose structured spatial pruning of connections between layers alongside temporal pruning by gradual timestep reduction while training for enhancing energy-efficiency of SNNs. Additionally, the network weights are quantized to perform further compression. First, the networks are trained till convergence. Then, to perform structured pruning, we find layerwise significant dimensions of the convolutional layers using Principal Component Analysis (PCA) of the average accumulated potential of the spiking neurons. A smaller network is then obtained with number of filters at each layer equal to that layer's significant dimension. Subsequently, this network is trained from scratch. We achieve up to 14X reduction in model size in this process by removing redundant spatial connections. In addition, we analyze the effect of spatial pruning on latency and computational requirements. Our results demonstrate that spatial pruning enables up to 2.5X latency reduction as well as 2-3X reduction in average number of spikes per inference compared to unpruned SNNs. To further enhance the energy-efficiency of SNNs, we perform temporal pruning on top of the spatially-pruned SNNs. The timestep of simulation is gradually reduced while training, so that the accuracy does not drop significantly. Notably, this allows us to obtain SNNs that can converge with $\sim$25 timesteps on CIFAR10 and $\sim$30 timesteps for CIFAR100, which is considerably lower compared to usual Poisson-coded SNNs \cite{hybrid}--\cite{rbodo}.
Note, here a single timestep denotes one full forward pass, in line with the definition of timestep used in previous SNN works \cite{hybrid}--\cite{rbodo}.
In addition, the spatio-temporally pruned SNNs provide 8-14X higher energy-efficiency compared to their ANN counterparts. Having performed the pruning steps, the weights of the network are quantized using weight-sharing via K-means clustering \cite{kmeans}. We utilize PCA-based structured pruning to reduce spatial connections unlike previous SNN pruning approaches which mostly focused on unstructured pruning. 
More importantly, we effectively utilize the time axis of SNNs to perform temporal pruning, leading to further energy efficiency. We perform our analysis on CIFAR10 and CIFAR100 using deep SNNs such as VGG9 and VGG11, respectively, to explore the various effects of compression beyond MNIST-like tasks. We also experiment with the robustness of the models against Gaussian noise, and find spatio-temporal pruning enhances robustness by 1-4\%. The main contributions of this work are summarized below -

\begin{itemize}

\item To the best of our knowledge, this is the first work that 
leverages the temporal axis of SNNs to perform pruning in order to reduce timesteps for computation in deep SNNs,  and as a result, enhances inference efficiency.

\item We incorporate PCA-based spatial pruning with temporal pruning, followed by quantization 
for designing efficient SNN architectures.

\item  We analyze the effect of spatial compression on temporal axis of SNNs, which reveals that spatial pruning leads to reduction in latency  and total spike count per inference.

\item The efficacy of the proposed methods is analyzed on CIFAR10 and CIFAR100 datasets with SNNs trained using the hybrid training method \cite{hybrid}. The resultant SNNs achieve 3-30X latency reduction compared to other state-of-the-art SNNs \cite{hybrid}--\cite{rbodo}, while needing 8-14X lesser compute energy compared to corresponding unpruned ANNs. 

\end{itemize}

The organization of the rest of this paper is as follows: the preliminaries of the SNN model and associated learning algorithms are presented in section II;
section III explains the methods of spatial and temporal pruning as well as quantization utilized in this work; section
IV comprises of the experimental setup, 
results, and related analysis, and the paper is concluded in Section V.

\section{Background}

\subsection{Spiking Neural Networks}

The Leaky Integrate and Fire (LIF) model
\cite{eliasmith} used in this work is described as-
\begin{equation}\label{eqn:1}
\tau_{m} \frac{dU}{dt}= -(U-U_{rest}) + RI,~~~~U\leq V_{th}
\end{equation}
where $U$ relates to the membrane potential, $I$ is the weighted sum of spike-inputs, $\tau_{m}$ indicates the time constant for membrane potential decay, $R$ represents membrane leakage path resistance, $V_{th}$ is the firing threshold and $U_{rest}$ is resting potential. The discretized version of Eqn.~\ref{eqn:1} is given as-
\begin{equation}\label{eqn:2}
u_i^t=\lambda u_i^{t-1}+\sum_{j} w_{ij}o_j^t-v_{th}o_i^{t-1},
\end{equation}
\begin{equation}\label{eqn:3}
o_i^{t-1}=\left\{
                \begin{array}{ll}
                  1,~~~if~u_i^{t-1}>v_{th}\\
                  0,~~~otherwise
                \end{array}
              \right.
\end{equation}
where $u$ is the membrane potential, subscripts $i$ and $j$ represent the post and pre-neuron, respectively, t denotes timestep, $\lambda$ is the leak constant= $e^{\frac{-1}{\tau_m}}$, $w_{ij}$ represents the weight of connection between the i-th and j-th neuron, $o$ is the output spike, and $v_{th}$ is the firing
threshold. As evident from Eqn.~\ref{eqn:2}, whenever $u$ crosses this threshold, it is reduced by the amount $v_{th}$, implementing a soft-reset. 

\subsection{Training Methodology}
The training scheme for deep SNNs can be broadly divided in to two categories - (i) ANN-SNN conversion: first an ANN is trained 
and then converted to an SNN by replacing the ReLU activation with IF/LIF neurons with threshold balancing \cite{diehl2015fast},\cite{sengupta2019going}, (ii) backpropagation from scratch: training SNNs from scratch using backpropagation is challenging due to the non-differentiability of the spike function. To overcome this, surrogate-gradient based learning \cite{neftci2019surrogate}
has been utilized where the derivative of the dirac-delta spike function is approximated. Another approach is hybrid training \cite{hybrid}, where a pre-trained ANN is first converted to SNN, then subsequent surrogate gradient learning is performed in the SNN domain. Conversion methods achieve high accuracy but suffer from high inference latency (2000-2500 timesteps). The latency is reduced in SNNs  trained  with  surrogate-gradients ($\sim$100-125 timesteps), but it needs to be improved further for edge deployment. In this work, we utilize the hybrid training method to demonstrate our method but the proposed scheme is applicable for both backpropagation from scratch and hybrid training approaches. 
For the hybrid method, we first train an iso-architecture ANN and copy the weights to the SNN for fine-tuning. Then to balance the layerwise neuronal thresholds, we select the 99.9 percentile
of the pre-activation distribution at each layer as its threshold as proposed in \cite{rbodo}. Next, we perform surrogate-gradient based learning, the details of which are depicted in Algorithm 1. The gradient of the spike function is approximated  using the linear surrogate-gradient \cite{bellec2018long}.

\begin{algorithm}[t]
   \caption{An iteration of spike-based backpropagation.}
   \label{alg:surrogate}
\begin{algorithmic}
   \STATE {\bfseries Input:} Pixel-based mini-batch of input ($X$) - target ($Y$) pairs, timesteps ($T$), number of layers ($L$), pre-trained ANN weights ($W$), membrane potential ($U$), membrane leak constant ($\lambda$), layer-wise firing thresholds ($V_{th}$)
   \STATE {\bfseries Initialize:} $U_l^t = 0,~\forall l = 1,...,L$
   
   \STATE //~{\bfseries Forward Phase}
   \FOR{$t \leftarrow 1$ {\bfseries to} $T$}
          
        \STATE //generate Poisson spike-inputs of a mini-batch data

        \STATE $O_1^t = Poisson(inputs);$ 
        \FOR{$l \leftarrow 2$ {\bfseries to} $L-1$}         
            \STATE //~membrane potential integrates weighted sum of spike-inputs
            \STATE $U_l^t~= U_l^{t-1} + W_{l}* O_{l-1}^t$
        \IF {$U_l^t > V_{th}$}
        \STATE //~if membrane potential exceeds $V_{th}$, a neuron fires a spike
            \STATE $O_{l}^t = 1,~U_l^t = U_l^t-V_{th}$
        \ELSE
        \STATE //~else, membrane potential decays exponentially
            \STATE $O_{l}^t  = 0,~U_l^t = \lambda * U_l^t$
        \ENDIF
        \ENDFOR    
        \STATE //~final layer neurons do not fire
        \STATE $U_{L}^t~= U_{L}^{t-1} + W_{L}* O_{L-1}^t$
   \ENDFOR
   \STATE //calculate loss, Loss=cross-entropy($U_{L}^T,Y$)
   \STATE //~{\bfseries Backward Phase}
   \FOR{$t \leftarrow T$ {\bfseries to} $1$}
       \FOR{$l \leftarrow L-1$ {\bfseries to} $1$}    
       \STATE //~evaluate partial derivatives of loss with respect to weight by unrolling the network over time
       \STATE $\triangle W_l^t = \frac{\partial Loss}{\partial O_l^t}\frac{\partial O_l^t}{\partial U_l^t}\frac{\partial U_l^t}{\partial W_l^t}$
       \ENDFOR
   \ENDFOR
   \end{algorithmic}
\vspace{-1mm}
\end{algorithm}

\section{Proposed compression methods}
In this section, we describe the proposed compression techniques in detail. Notably, the spatial and temporal pruning methods can be applied independently or in conjunction.

\subsection{PCA-based Spatial Pruning}
\begin{figure*}[t]
\centering
\includegraphics[width=\linewidth, height=2.7in]{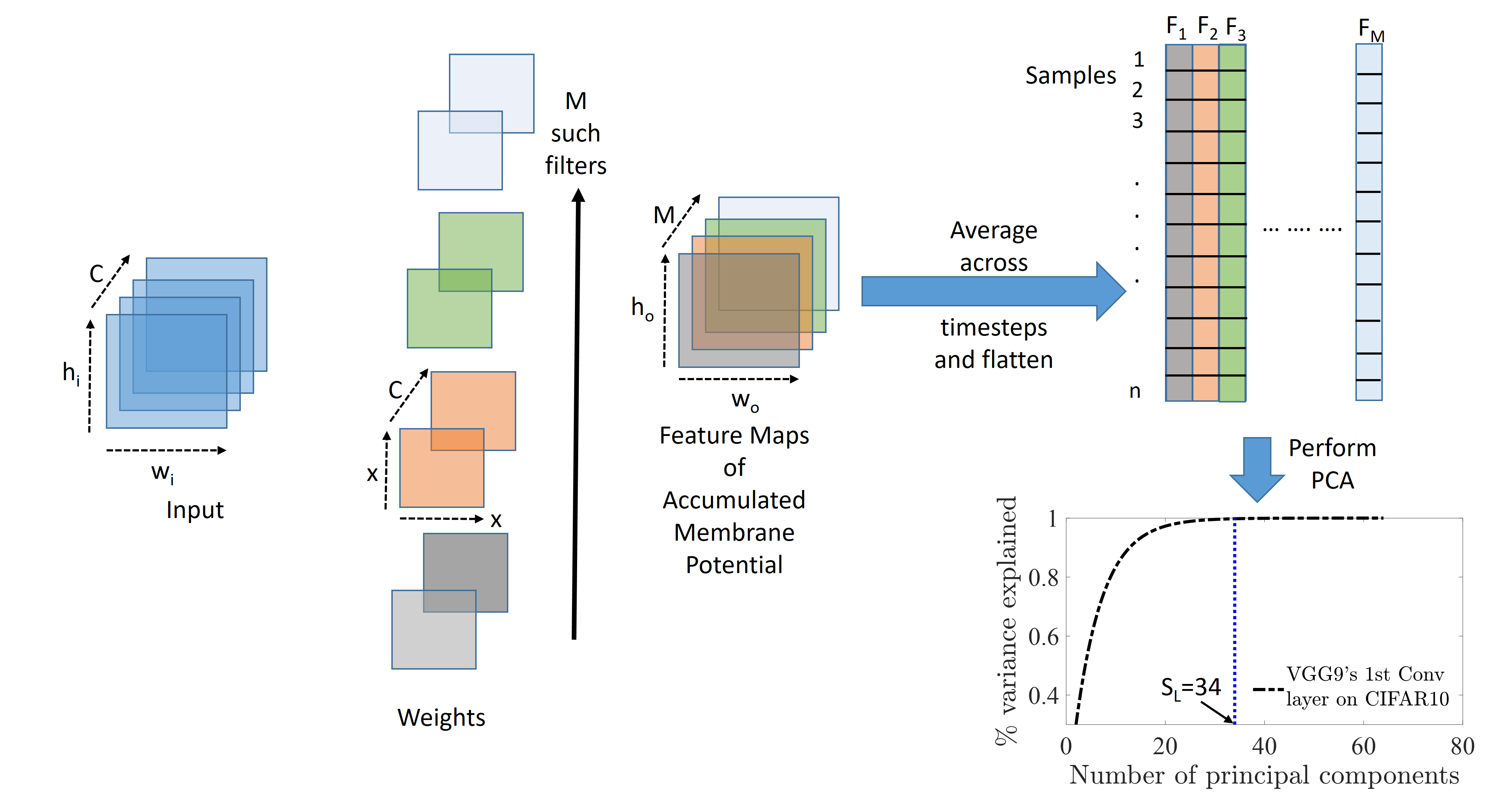}
\vspace{-3mm}
\caption{Schematic of PCA-based pruning of SNNs, here $S_L$ denotes number~of significant dimensions explaining
99.9\% of the total variance in data.}
\label{fig1}
\vspace{-3mm}
\end{figure*}
Pruning methods aim to compress networks post training in order to remove redundancy by either pruning individual
weights or removing entire filters. Pruning connections based on magnitude thresholding results in unstructured sparsity that is difficult to leverage in
hardware. Hence, in this paper, we focus on obtaining structured sparsity in trained SNNs using principal component analysis (PCA) following the method proposed in \cite{pca} for ANNs. In CNNs, there exists high correlation among many filters within the same layer,
thus we can reduce dimensionality of layers without causing significant hit to accuracy. Reference \cite{pca} proposed identifying such correlations by applying PCA on the activation maps
generated by the filters in ANNs, and we adopt this approach for the case of SNNs. The goal is to determine the number~of significant dimensions at each conv layer to find the least number~of features required for explaining
99.9\% of the total variance in input. The information propagation for SNNs is embedded in the spiking activity of the neurons. If multiple neurons spike in a coordinated manner, there is redundancy among them and they can be reduced to a smaller set of neurons with modified weights. In order to detect redundancy between
filters, we use the pre-spike activations (accumulated membrane potential) as
instances of filter activity that serve as feature value inputs to PCA. Note, in this case, we utilize the average membrane potential accumulated at the neurons of the corresponding layer using incoming spike inputs over all timesteps. A schematic of the proposed PCA-based pruning is depicted in Fig.~\ref{fig1}. Here, $C$ ($M$)
denote the number of input (output) channels for a convolutional layer and $h_i;w_i (h_o;w_o)$ denote the height and width
of the input (output) feature maps. First, the SNN is trained till convergence with an initial configuration, then a forward pass is performed to find the redundancy of filters. Let, $B_L$ denotes the activation matrix obtained as the output
of a forward pass, here $L$ refers to the layer index. The top-left input patch is convolved with the first filter to give the top left pixel of the output activation
map. The convolution of the same patch with all M filters provides a vector $\in \mathbb{R}^{1\times1\times M}$, which counts as one sample of M parameters, where each parameter represents the activity
of a filter. Similarly, the next filter activity sample is obtained by sliding the convolution kernel to the next input patch. This process is detailed in \cite{pca}. Let, $B_L\in \mathbb{R}^{b\times h_o\times w_o\times M}$, where b is the mini-batch
size, so we collect $b\times h_o\times w_o$ samples in one forward
pass, with $M$ parameters each. We perform the convolution at each timestep and the output is accumulated as membrane potential at the neurons. Then, we average the accumulated membrane potential over all timesteps which acts as a measure of the spike-rate of the neurons. Then this averaged $B_{L,avg}$ is flattened as $B_{L,avg}\in \mathbb{R}^{b\times h_o\times w_o\times M} \to C_L\in \mathbb{R}^{S\times M}$, with $S=b\times h_o\times w_o$. This matrix $C_L$ is given as input to PCA where Singular Value Decomposition (SVD) is performed on the
matrix $C_L^TC_L$. Using this operation, we find significant dimensionality
of the compressed space of filters, defined as the number of
uncorrelated filters that can explain 99.9\% of the variance of features, as shown in Fig.~\ref{fig1} (the curve in the figure corresponds to 1st conv layer of VGG9 on CIFAR10). Thus, the layerwise width (number~of filters) of the network is optimized as suggested in \cite{pca}. Note, this dimensionality reduction can be performed in parallel for all
layers which means we get the significant dimension at all layers concurrently with just a few forward passes. Furthermore, the depth (number~of layers) is also reduced following a heuristic proposed in \cite{pca}, where we get rid of the layers where number~of significant dimensions  decrease  compared to the preceding layer. 
Once the PCA-pruned configuration (where the significant dimensions determine the width at each layer) is determined, we initialize an ANN with this compressed architecture and perform the whole hybrid training pipeline to obtain the pruned SNN.

\subsection{Temporal Pruning While Training}
The spatial pruning compresses the networks spatially. However, SNNs consist of a temporal axis, whereby the information is spread across multiple timesteps. To leverage the time dimension and reduce latency, we propose gradual pruning of timesteps while training. The method is described in Algorithm 2. First, the SNN is trained till convergence with maximum number~of timesteps that yields highest accuracy. Then, we start reducing the simulation timesteps by $v$ at each step and keep training the SNN for few epochs to regain accuracy. Here, the number~of timesteps to curtail at a single pruning iteration ($v$) as well as number~of epochs to retrain at each step are hyperparameters. We perform this timestep pruning anticipating it will lead to reduction in latency as well as enhance energy-efficiency, which is validated in section IV. However, there is an inherent trade-off between accuracy and latency in SNNs. 
Hence, the validation accuracy is checked before each timestep pruning iteration and the stopping criterion is based on the lowest acceptable validation accuracy. This temporal pruning scheme is motivated by spatial pruning methods where a network is first trained, then based on some importance measure, insignificant connections are pruned away. Likewise, we first train the SNN with higher number~of timesteps and then gradually perform  timestep reduction. Notably, this temporal pruning step can be performed either in standalone fashion or together with the structured spatial pruning step described previously. Though the pruning scheme described above increases the training overhead slightly, 
the goal here is to reduce inference latency and enhance inference efficiency.

\begin{algorithm}[t]
   \caption{Pseudo-code for temporal pruning while training.}
   \label{alg:tempprun}
\begin{algorithmic}
   \STATE {\bfseries Input:}  Trained SNN with total number of timesteps ($T$), desired reduced timesteps ($T_r$), timesteps reduced in single step ($v$), number~of epochs to train between timestep reductions ($e$), lowest acceptable validation accuracy ($A_{min}$)
   \STATE {\bfseries Initialize:} $T_r = T-v$, validation accuracy, $A_{tpruned}=$ validation accuracy of network trained with $T$ timesteps
   
   \WHILE{$A_{tpruned}>A_{min}$}
   
   \STATE //~{\bfseries Training Phase}
   \FOR{$epoch \leftarrow 1$ {\bfseries to} $e$}
          
        \STATE //Train network with $T_r$ timesteps using algorithm 1

   \ENDFOR
   \STATE //~{\bfseries Validation Phase}
   \STATE //~Find validation accuracy
   \STATE $A_{tpruned}$= Accuracy(network trained in previous step)
   
   \STATE //~ Timestep reduction
   \STATE $T_r = T_r-v$
   \ENDWHILE
   
   \end{algorithmic}
\end{algorithm}
 
\subsection{Weight Quantization}
In addition to the pruning methods, weight quantization can be performed to further compress the SNNs. The activations of SNNs are binary, so if the weights can be represented with reduced bit-precision without accuracy loss, it can reduce the storage requirements significantly. Here, we adopt weight-sharing method as proposed in \cite{deepcompression} to limit the number of effective weights. K-means clustering \cite{kmeans} is used to cluster similar weights for each layer of a trained network where
all the weights belonging to the same cluster share the same weight. Notably, we perform this clustering both for the conv as well as the FC-layers. With $z$ clusters, only $log_2(z)$ bits are needed to encode the index of weights. For a network with $p$ connections (each represented with $b$ bits), the compression rate in case of $z$ shared weights becomes \cite{deepcompression}:
\begin{equation}\label{eqn:4}
r= \frac{pb}{plog_2(z)+zb}
\end{equation}
We would like to mention that this quantization step is orthogonal to previous PCA-based pruning and timestep reduction approaches. Again, there is an inherent trade-off between bit-precision and accuracy. Hence, it provides a knob to obtain an optimized configuration with lowest bit-precision possible with minimal accuracy drop.
\begin{table}[t]

\caption{PCA-based filter dimensionality reduction}
\begin{center}
\begin{tabular}{|c|c|c|c|}
\hline
VGG9& Initial Dim.& \multicolumn{2}{|c|}{64,~128,~128,~256,~256,~512,~512} \\
\cline{2-4} 
CIFAR10 & Significant Dim. & \multicolumn{2}{|c|}{34,~118,~123,~250,~244,~496,~503} \\
\cline{2-4} 
 & Final Dim. & \multicolumn{2}{|c|}{34,~118,~123,~250,~496,~503} \\
\hline
VGG11& Initial Dim.& \multicolumn{2}{|c|}{64,~128,~256,~512,~512,~512,~512,~512} \\
\cline{2-4} 
CIFAR100 & Significant Dim. & \multicolumn{2}{|c|}{48,~114,~241,~497,~496,~484,~497,~500} \\
\cline{2-4} 
 & Final Dim. & \multicolumn{2}{|c|}{48,~114,~241,~497,~497,~500} \\
\hline

\end{tabular}
\label{tab1}
\end{center}
\end{table}
\section{Experiments and Results}

\subsection{Experimental Setup}
The experiments are performed on VGG9 for CIFAR10 dataset and on VGG11 for CIFAR100. All results are obtained using the hybrid training approach \cite{hybrid}. For all cases, standard data augmentation
techniques are  followed such as padding by 4 pixels on each side, and a $32\times32$ random cropping from the padded image or its horizontally flipped version (with 0.5 probability of flipping). The testing is done using the original $32\times32$ images. Both training and testing data are normalized using 0.5 as both mean and standard deviation for all channels. While training the ANNs for hybrid training method, we use cross-entropy loss with stochastic gradient descent optimization (weight decay=0.0001, momentum=0.9) with an initial learning rate of 0.1,
which is divided by 10 every 100-th epoch. The ANNs do not have bias terms and batch-normalization is
not used, rather dropout is used as regularizer and the
dropout mask is kept constant across all timesteps in the SNN domain. Furthermore, to perform pooling between layers, average-pooling
is used instead of max-pooling to avoid
information loss in SNNs as per \cite{diehl2015fast}. Poisson rate-coding is used to convert static images into spike inputs over time.
  Upon ANN-SNN conversion, SNNs are trained for 20-30 epochs with cross-entropy loss
and adam optimizer (weight decay=0.0005). Initial learning rate is kept at 0.0001,
which is halved every 5-th epoch. 
\begin{figure}[t]
\centering
\includegraphics[width=\linewidth, height=3.1in]{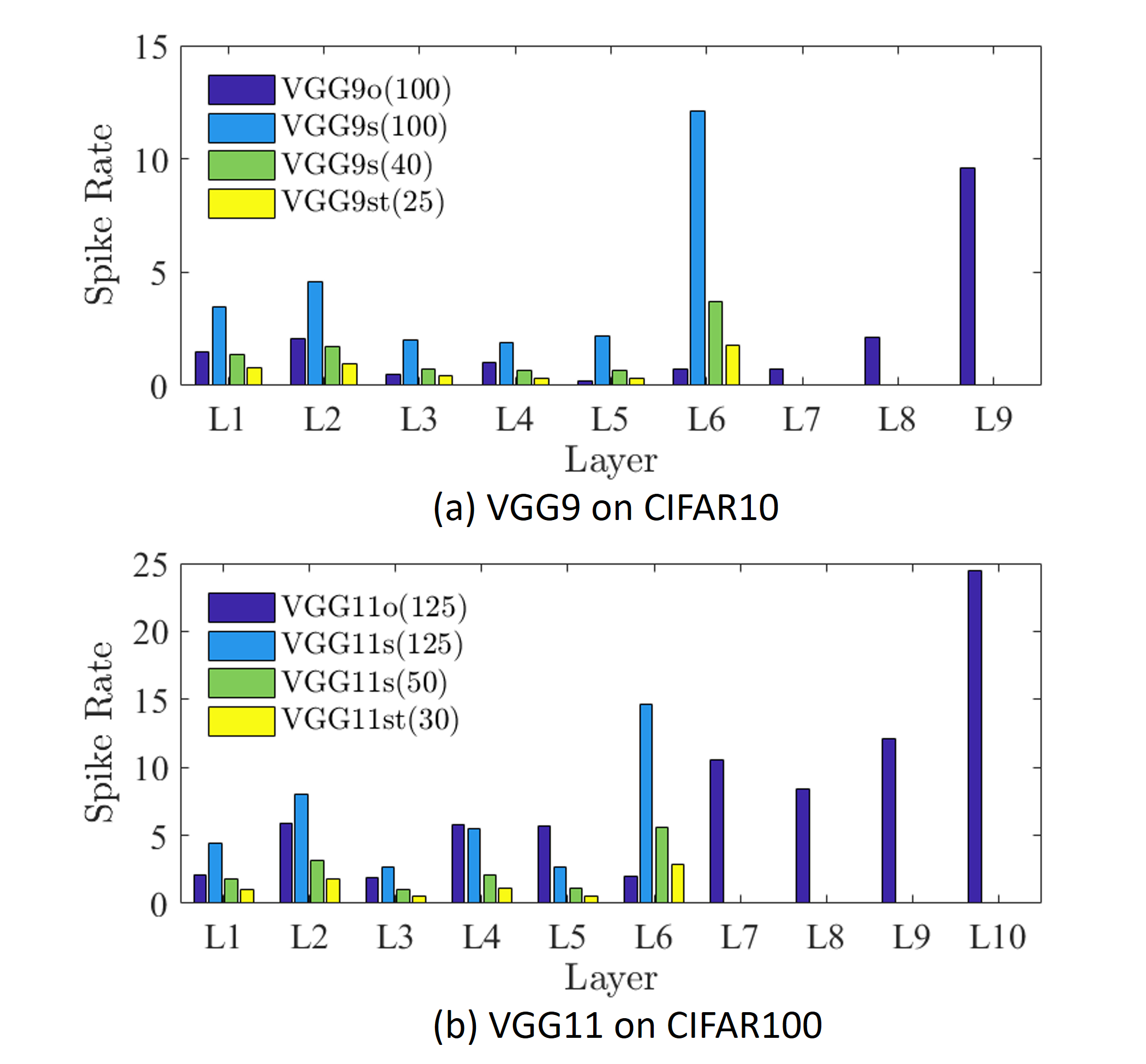}

\caption{Layerwise spike rate for (a) VGG9 on CIFAR10, (b) VGG11 on CIFAR100, here VGGmo, VGGms and VGGmst denote original unpruned, spatially pruned and spatio-temporally pruned network, respectively (the corresponding timesteps are denoted in parenthesis). Li denotes i-th layer. The later layers contain only VGGmo results due to depth reduction in VGGms and VGGmst networks through spatial pruning.}
\label{fig2}
\vspace{-6mm}
\end{figure}
\subsection{Effect of PCA-based Spatial Pruning}
First, we analyze the effect of PCA-based spatial pruning of conv layers of SNN using the method described in section III, subsection A. 
The layerwise significant dimensions obtained using PCA of the average accumulated membrane potential are given in Table \ref{tab1}. As can be seen, our method successfully identifies redundancies in the filters across the layers. Having determined the minimum number~of filters which captures most of the data variation in each layer, we initialized a new network with the reduced dimensions, also the 3 fully-connected layers of VGG networks are reduced to a single layer as per \cite{pca}. For the rest of this paper, we refer to the original unpruned, PCA-based spatially compressed, spatio-temporally compressed and spatio-temporally compressed as well as quantized VGGm networks as `VGGmo', `VGGms', `VGGmst' and `VGGmstq', respectively (here m=9 or 11). The accuracies obtained on CIFAR10 with VGG9o and VGG9s networks are 90.1\% and 90.02\% respectively, with 100 timesteps. For CIFAR100, the accuracies obtained with VGG11o and VGG11s networks are 68.1\% and 68\% respectively, with 125 timesteps. So, there is $<$0.1\% drop in accuracy with the spatial pruning of filters. In this case, the depths of the networks are compressed by deleting the layers which do not increase the significant dimensionality compared to previous layer following the heuristic described in \cite{pca}. We reduce the depth by 1 and 2 layers for VGG9 and VGG11, respectively. With this spatial pruning step, the model sizes and number~of parameters are reduced. The model sizes for VGG9o and VGG9s are 629.7MB and 44.5MB, respectively. For, VGG11o, and VGG11s, the model sizes are 706.4MB and 75.9 MB, respectively. In terms of parameters, compared to VGG9o, VGG9s has $0.07$X parameters. Again, VGG11s has $0.107$X  parameters compared to VGG11o. These results are shown in Table II. We also investigate the effect of the leak constant ($\lambda$) on the layerwise significant dimensions, but observe that the filter space dimensionality has no dependence on it. So, we choose $\lambda$ to be 0.9901 for all simulations, since this value of leak provides optimized performance for SNNs in terms of robustness and spiking activity as per \cite{leak}. 
\vspace{-1mm}
\begin{figure}[t]
\centering
\includegraphics[width=\linewidth]{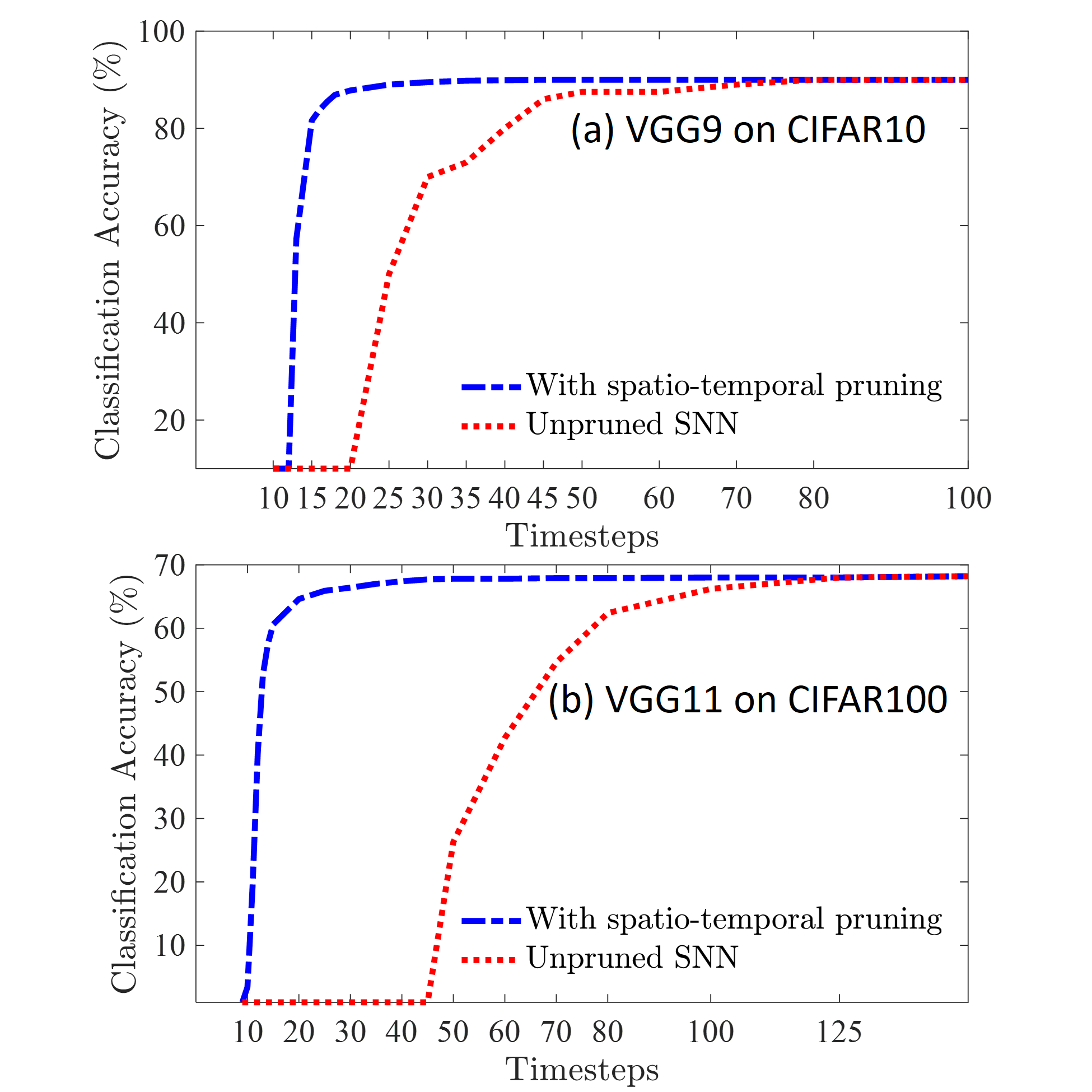}
\caption{Accuracy versus latency curve for Poisson-encoded SNNs.}
\label{fig3}
\vspace{-4mm}
\end{figure}
\subsection{Effect of Spatial Pruning on Latency and Compute Requirements}

Next, we investigate the effect of spatial pruning on latency and average number of spikes per inference. This analysis holds importance since the energy-efficiency of SNNs is directly related to the number~of operations performed, which is determined by the total number of spikes per inference across all layers. In this regard, we compare the cumulative number of spikes per inference for all layers averaged over all images (termed as average cumulative spike-count per inference, or ASCI) between the unpruned and pruned networks. Interestingly, ASCI for VGG9s becomes 1.8X compared to VGG9o. To counter this, we next investigate if the spatially-pruned networks can converge with lower timesteps. The intuition is that since each timestep incurs more spikes to the previous cumulative spike count, timestep reduction might lead to decrease in ASCI. Our results validate that the spatially pruned networks indeed enable convergence at $\sim$2.5X lower latency with negligible drop in accuracy. As shown in Table II, VGG9s(40) and VGG11s(50) have $\leq0.3\%$ accuracy drop compared to VGG9o(100) and VGG11o(125), respectively (the numbers in parenthesis denote timestep). The reason behind this latency reduction with spatial pruning is depth reduction. Deep SNNs need to propagate spikes in a layerwise manner from the input to the final layer and each layer adds to the latency requirement since the potential needs to be accumulated before the current layer can propagate spikes to next layer. Since some layers are removed through spatial pruning, the resultant SNNs can converge faster. On the other hand, since the spatially pruned networks reach convergence faster than the unpruned networks, simulating them for equal number of timesteps causes redundant spikes in the pruned networks. Hence, VGG9s(100) has higher ASCI than VGG9o(100). However, the ASCI of VGG9s(40) is 0.65X compared to VGG9o(100), similarly, the ASCI of VGG11s(50) is 0.32X compared to VGG11o(125). To further analyze the spiking activity of various networks, the layerwise spike rates of each network are  shown in Fig.~\ref{fig2}.  The spike rate at a layer L of a network is defined as \cite{hybrid,dctsnn}-
\begin{equation}\label{eqn:5}
\text{Spike Rate$_L$}=\frac{\#\text{Spikes over all timesteps in layer L }}{\# \text{Neurons in layer L} }
\end{equation}
From Fig.~\ref{fig2}, we again notice similar pattern with VGG9s(100) and VGG11s(125) showing higher layerwise spiking activity compared to their unpruned counterparts. However, this issue is resolved with VGG9s(40) and VGG11(50). The average spike rates for VGG9o(100), VGG9s(100) and VGG9s(40) are 1.38, 3.45 and 1.31, respectively. Again, for VGG11o(125) VGG11s(125) and VGG11s(50), the average spike rates are 4.21, 4.67 and 1.84, respectively. So, spatial pruning enables inference with lower latency as well as lower compute requirements (by decreasing ASCI and spike rate), thereby improving energy-efficiency. We report results for VGG9s and VGG11s upto 40 and 50 timsteps, respectively, below which the networks do not converge well. To further enhance SNN performance by reducing latency even more through leveraging the time axis of SNNs, we perform temporal pruning, which is the focus of our next discussion.
\begin{table*}[t]

\caption{Comparison among unpruned, spatially pruned, spatially and temporally pruned and spatially and temporally pruned as well as quantized networks, here ASCI denotes average cumulative spike count per inference}
\begin{center}
\begin{tabular}{|c|c|c|c|c|c|c|c|c|c|c|}
\hline
Network & VGG9o & VGG9s& VGG9s& VGG9st& VGG9stq&VGG11o & VGG11s& VGG11s& VGG11st& VGG11stq\\
\hline
Accuracy(\%) &90.1 &90.02 &89.9&89.04&88.6 &68.1&68 &67.8&66.4 &66.2  \\
\hline
Timestep &100 &100 &40&25 &25 &125&125 &50&30 &30  \\
\hline
\#Params &1X &0.07X &0.07X&0.07X&0.07X &1X&0.107X
 &0.107X&0.107X &0.107X  \\
\hline
ASCI &1X &1.8X &0.65X&0.35X&0.35X &1X&0.82X
 &0.32X&0.18X &0.18X  \\
\hline
Bit Precision &32 &32 &32&32&5 &32&32&32&32&5  \\
\hline
\end{tabular}
\label{tab2}
\end{center}
\vspace{-6mm}
\end{table*}
\subsection{Analysis of Temporal Pruning}
We perform gradual temporal pruning while training as discussed in section III, subsection B. 
The effect of temporal pruning on spatially pruned networks is shown in Fig.~\ref{fig3}. Interestingly, for both VGG9 and VGG11, the proposed method enables us to keep reducing timesteps to much lower limits compared to what would be possible without any spatio-temporal pruning. In the regions where usual training method fails to converge ($\leq30$ timesteps for VGG9 and $\leq50$ timesteps for VGG11), our pruned networks retain near maximum performance. However, one drawback is that to perform such temporal pruning, first we need to train the SNN with higher timesteps and then gradually reduce timesteps while training, which increases the training effort. For our experiments, at each timestep reduction iteration, we reduce timestep by 1, and this configuration is retrained for 1 epoch to regain accuracy. Note,  without the pre-training with higher timesteps, we did not succeed in obtaining a network with reduced latency directly. For example, for VGG9 on CIFAR10, starting with a network
trained on 100 timesteps, we perform spatio-temporal pruning to reach a network which achieves 81.63\% accuracy with just 15 timesteps. But a network trained from scratch with 15 timesteps does not converge at all. This validates the efficacy of our method in training networks with extremely low latency and high efficiency requirements, suitable for real-time applications and edge deployment. This method of first training a larger network and then obtaining a smaller sub-network (in the temporal axis) is similar to what is followed in spatial pruning. This approach can be thought of as a kind of simulated annealing, where the pre-training with higher timesteps provides a suitable initialization for subsequent latency reduction. Another perspective is to think of this method as a form of distillation process, where the initial network with higher timesteps works as the teacher network, and the student network with reduced latency is learnt from it. However, in our case, the student configuration is not pre-defined, rather it gradually evolves from the teacher at each pruning iteration. Again, distillation works better when the mismatch of capacity between student and teacher networks is minimized \cite{distill2}. In our case, this is satisfied implicitly since at each pruning iteration, the student learns from the teacher trained at previous iteration, so the student and teacher differ by latency of single timestep. The accuracies and ASCI results of the spatio-temporally pruned networks are given in Table \ref{tab2}. As can be seen, for VGG9st and VGG11st, the timesteps required and ASCI can be reduced significantly, albeit with little drop in accuracy. Furthermore, the layerwise spike rates for the VGG9st and VGG11st networks are shown in Fig.~\ref{fig2}, which demonstrates that the spike rate at each layer are reduced noticeably compared to unpruned configurations. 
\begin{figure}[t]
\centering
\includegraphics[width=.8\linewidth]{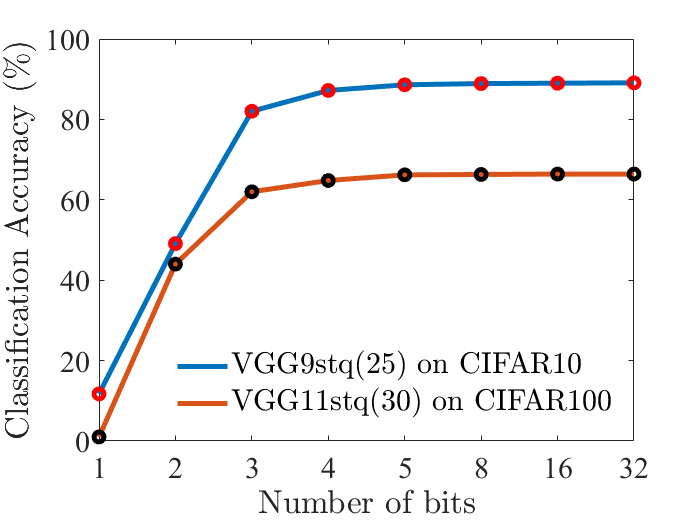}
\caption{Effect of weight quantization on accuracy.}
\label{fig4}
\vspace{-4mm}
\end{figure}

\subsection{Quantization Results}

The primary focus of this paper is the spatio-temporal pruning methods described so far, however we also explore the feasibility of performing quantization on top of the pruned networks. The weight sharing quantization adopted in this work has been described in section III, subsection C. Here, the quantization is applied on top of spatio-temporal pruning and the quantized networks are denoted as VGG9stq and VGG11stq. The results are shown in Fig.~\ref{fig4}. For both cases, up to 5 bit precision, the accuracy drop is negligible, which is consistent with results reported in \cite{admm}. Beyond that, at 2 bit quantization level, there is significant accuracy degradation and at 1 bit, the networks fail to converge. This shows that quantization methods can be easily applied up to 5 bits together with the proposed pruning schemes to further enhance the SNN compression performance without sacrificing accuracy. Note, we perform weight clustering in a layerwise manner at same bit level for all layers. For 5-bit quantization, 
model sizes for VGG9stq and VGG11stq are only 0.02X and 0.036X, respectively of VGG9o and VGG11o. So, the storage requirements for these SNNs are significantly reduced using the proposed compression pipeline.

\subsection{Computational Efficiency}
SNNs derive their energy-efficiency by replacing the floating-point (FP) multiply and accumulate (MAC) operations by FP additions. In 45nm
CMOS technology, an addition ($0.9pJ$) is $5.1\times$ less costly than a MAC ($4.6pJ$) \cite{horowitz20141} . We calculate the layerwise number of operations (\#ops) of the parent unpruned ANN using-
\vspace{-2mm}
\begin{equation*}
\begin{split}
\# \text{ANN}_{\text{ops}} = \left\{
                \begin{array}{ll}
                  k_w\times k_h\times c_{in}\times h_{out} \times w_{out} \times c_{out},\\~~~~~~~~~~~~~~~~~~~~~~~~~~~~~~~~~\text{Conv layer}\\
                  n_{in}\times n_{out},~~\text{Linear layer}
                \end{array}
              \right.
\end{split}
\end{equation*} 
 where $k_w (k_h)$ denote filter width (height), $c_{in} (c_{out})$ is number of input (output) channels,
$h_{out} (w_{out})$ is the height (width) of the output feature map, and $n_{in} (n_{out})$ is the number of input
(output) nodes. The \#ops in layer L of an SNN is related to \#ops of an iso-architecture ANN 
by that layer's spike-rate-
\begin{equation}
\#\text{SNN}_{\text{ops, L}} = \text{spike rate}_\text{L} \times \#\text{ANN}_{\text{ops, L}}, 
\end{equation}
The layerwise spike rates for CIFAR10 and CIFAR100 using the proposed pruning methods are shown in Fig.~\ref{fig2}. We calculate the layerwise \#ops of the pruned SNNs using Eqn.~6. Then the energy benefit of the SNNs over the unpruned ANN ($\alpha = \frac{\text{E}_{\text{Unpruned ANN}}}{\text{E}_{\text{Compressed SNN}}}$) is computed as, 
\vspace{-1.5mm}
\begin{equation}
    \alpha=
    \frac{\sum_{L} \#\text{Unpruned ANN}_{\text{ops,L}}*4.6}{\sum_{L} \#\text{Compressed SNN}_{\text{ops,L}}*0.9}
\end{equation} 
We find that the VGG9st(25) and VGG11st(30) networks are $14.64\times$ and $8.43\times$ energy-efficient, respectively, compared to their corresponding unpruned ANNs. Also, in comparison to VGG9o(100) and VGG11o(125), VGG9st(25) and VGG11st(30) provide $5.5\times$ and $7.7\times$ energy-efficiency, respectively. Note, this evaluation of efficiency excludes the  cost  of  memory  access, since it is hardware architecture and system configuration dependent.
\begin{figure}[t]
\centering
\includegraphics[width=.8\linewidth]{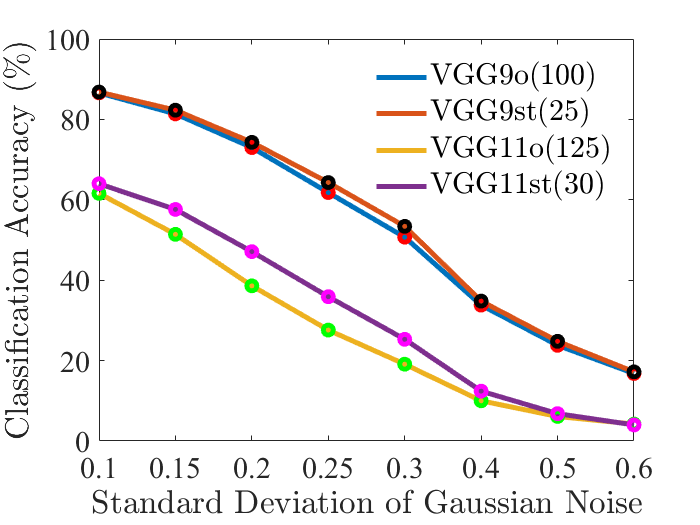}
\caption{Effect of pruning on robustness to random Gaussian noise.}
\label{fig5}
\vspace{-4mm}
\end{figure}
\begin{figure}[t]
\centering
\includegraphics[width=\linewidth, height=4in]{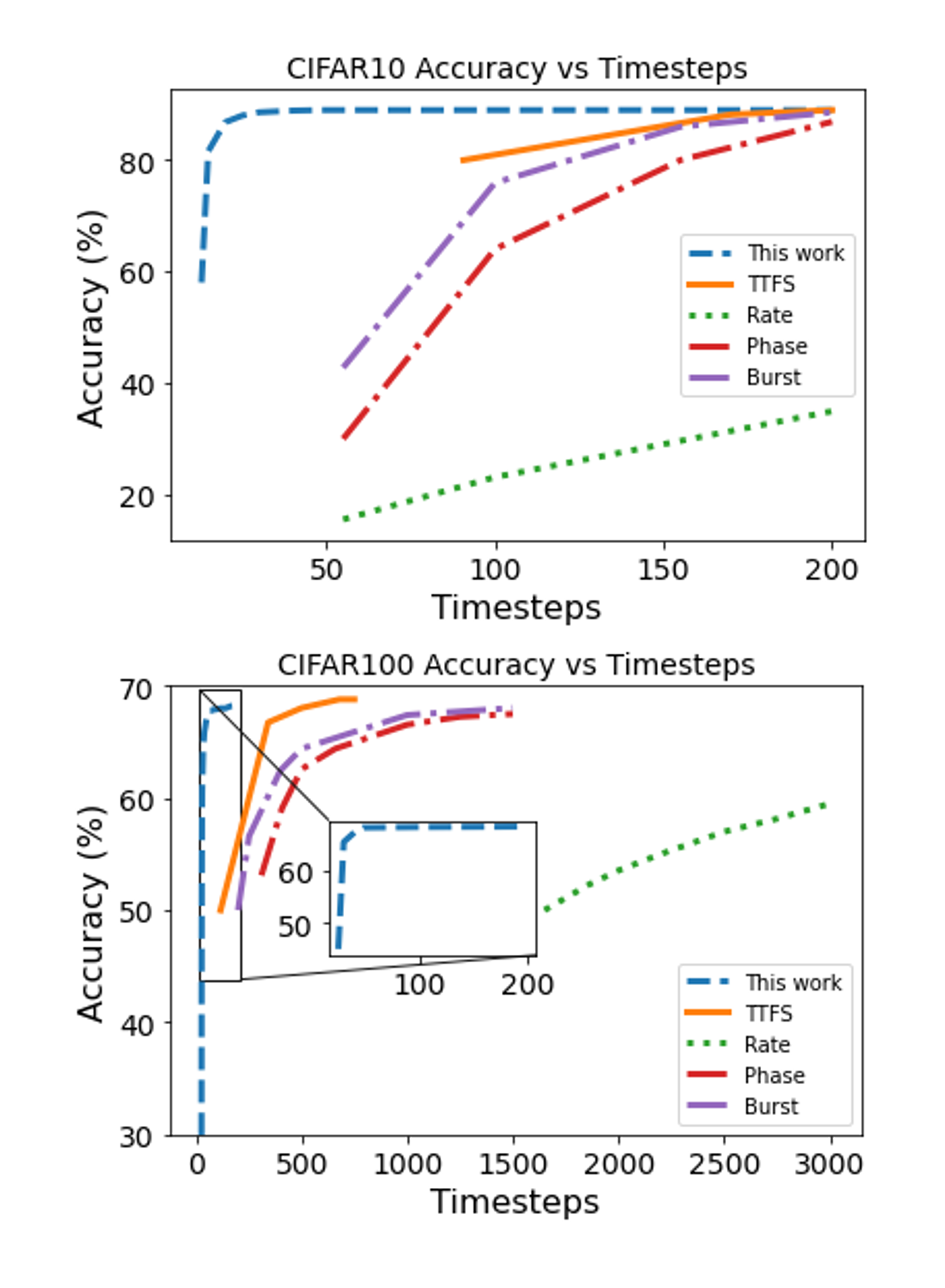}
\caption{Accuracy versus latency curve for various recent algorithms, the values for time-to-first-spike (TTFS) \cite{ttfs}, Phase \cite{phase}, Burst \cite{burst} and Rate \cite{rbodo} have been adopted from \cite{ttfs}.}
\label{fig6}
\vspace{-4mm}
\end{figure}
\subsection{Robustness to Noise}
Next, we analyze the robustness of the spatio-temporally pruned nets to Gaussian noise and the results are displayed in Fig.~\ref{fig5}. Here, the noise is applied to the analog-valued pixels before converting them to spikes. Compared to VGG9o and VGG11o, VGG9st and VGG11st provide 1.3\% and 4.3\% enhanced robustness respectively, averaged across different intensities of noise (determined by noise standard deviation). Similar noise robustness has been observed for spatially pruned networks in \cite{nettrim}, where it is attributed to the implicit regularization obtained by 
eliminating redundant connections. 
Our results are also consistent with \cite{saima}, where the authors reported enhanced robustness to adversarial noise for networks trained with lesser timesteps.

\subsection{Comparison with Prior Art}

In this section, we compare the performance of our spatio-temporally pruned networks with other recent algorithms in terms of providing low-latency solution for SNNs. The accuracy versus latency curves for CIFAR10 and CIFAR100 are shown in Fig.~\ref{fig6} top and bottom, respectively. Compared to time-to-first-spike (TTFS) coding \cite{ttfs}, Phase coding \cite{phase}, Burst coding \cite{burst} and other Rate \cite{rbodo} coding approaches, our proposed method can converge at much lower latency for both datasets. The drawback is that it suffers from a small accuracy drop. However, the range of timesteps where our method can reach near convergence, other algorithms fail to converge. This validates the efficacy of the proposed compression framework in obtaining extremely low-latency, low-power SNNs suitable for edge applications. In addition, we also provide comparison of this work with some other reported results in Table \ref{tab3}. As can be seen, compared to other works, the spatial and temporal pruning approach proposed here enables SNNs to obtain higher performance in comparatively lesser timesteps. Moreover, this latency reduction also translates to higher energy-efficiency, as we have shown before.
\begin{table}[t]

\caption{Comparison of the proposed method to other reported results}
\begin{center}
\begin{tabular}{|c|c|c|c|}
\hline
Reference & Dataset & Accuracy (\%) & Timestep   \\

\hline
Hunsberger \textit{et al.~}\cite{eliasmith} & CIFAR10 & 82.95 &6000  \\
\hline
Cao \textit{et al.~}\cite{cao} & CIFAR10 & 77.43 &400 \\
\hline

Wu \textit{et al.~}\cite{wu18} & CIFAR10 & 50.7 &30  \\
\hline
Kushawaha \textit{et al.~}\cite{distill} & CIFAR10 & 45.98 & not reported  \\
\hline
Srinivasan \textit{et al.~}\cite{gopal} & CIFAR10 & 66.23 & 25  \\
\hline

\textbf{This work} & CIFAR10 & 89.04 &25  \\
\hline
Lu \textit{et al.~}\cite{lu} & CIFAR100 & 63.2 &62  \\
\hline
Rathi \textit{et al.~}\cite{hybrid} & CIFAR100 & 67.9 &125  \\
\hline
Sharmin \textit{et al.~}\cite{saima} & CIFAR100 & 64.1 &200  \\
\hline

Kundu \textit{et al.~}\cite{spike-thrift} & CIFAR100 & 64.98 &120  \\
\hline
\textbf{This work} & CIFAR100 & 66.4 &30  \\
\hline
\end{tabular}
\label{tab3}
\end{center}
\vspace{-6mm}
\end{table}

\section{Conclusion}

SNNs can potentially provide an attractive energy-efficient alternative to ANNs. However, it is pivotal to address the issue of high inference latency for rate-coded SNNs. Merging usual compression approaches used in ANNs (such as pruning and quantization) with SNNs could enhance their energy-efficiency benefits. To that end, we propose spatial and temporal pruning and quantization for low-latency, highly efficient SNNs. We validate our results using VGG networks on CIFAR10 and CIFAR100. PCA-based spatial pruning compresses model size and reduces latency and compute requirements. In the proposed scheme, spatial pruning is followed by temporal pruning, which further reduces latency and improves energy-efficiency. Additional model compression is achieved using weight sharing quantization. This work applies compression techniques traditionally used for ANNs in the realm of SNNs and proposes to utilize the time axis of SNNs effectively to perform holistic spatio-temporal compression. The associated trade-off is slight accuracy degradation, but allows SNNs to infer with 8-14X  energy efficiency  compared  to  unpruned ANNs, while occupying 10-14X lesser memory. We believe this approach provides a pathway towards finding suitable solutions to train SNNs with very low energy and latency requirements, which is crucial for edge applications. Future works include exploring other pruning and quantization methods for SNN compression.

\section*{Acknowledgment}

The work was supported in part by, Center for Brain-inspired Computing (C-
BRIC), a DARPA sponsored JUMP center, Semiconductor Research Corporation, National Science Foundation, the DoD Vannevar Bush
Fellowship and U.S. Army Research Laboratory.


\begin{thebibliography}{00}
\bibitem{b1} A. Krizhevsky,  I.  Sutskever,  and  G. E.  Hinton, ``Imagenet  classification  with  deep convolutional neural networks,'' In Advances in neural information processing systems, pp. 1097–1105, 2012. 
\bibitem{b2} A. Vaswani, N. Shazeer, N. Parmar, J. Uszkoreit, L. Jones, A. N. Gomez,
Ł. Kaiser, and I. Polosukhin, “Attention is all you need,” in Advances
in Neural Information Processing Systems, pp. 5998–6008, 2017.
\bibitem{li2016evaluating} D. Li,  X. Chen,  M. Becchi,  and  Z. Zong, ``Evaluating  the  energy  efficiency  of deep convolutional neural networks on cpus and gpus,''  In 2016 IEEE international conferences on big data and cloud computing (BDCloud),  social computing and networking (SocialCom), sustainable  computing  and  communications  (SustainCom) (BDCloud-SocialCom-SustainCom), pp. 477–484, IEEE, 2016
\bibitem{lowrank} A. Novikov, D. Podoprikhin, A. Osokin, and D. P. Vetrov, “Tensorizing
neural networks,” in Advances in Neural Information Processing
Systems, pp. 442–450, 2015.

\bibitem{han2015} S. Han, J. Pool, J. Tran, and W. Dally, “Learning both weights and connections
for efficient neural network,” in Advances in neural information
processing systems, pp. 1135–1143, 2015.

\bibitem{wen2016} W. Wen, C. Wu, Y. Wang, Y. Chen, and H. Li, “Learning structured
sparsity in deep neural networks,” in Advances in Neural Information
Processing Systems, pp. 2074–2082, 2016.

\bibitem{binarynet}M. Courbariaux, I. Hubara, D. Soudry, R. El-Yaniv, and Y. Bengio,
“Binarized neural networks: Training deep neural networks with
weights and activations constrained to +1 or -1,” arXiv preprint
arXiv:1602.02830, 2016.


\bibitem{haq}K. Wang, Z. Liu, Y. Lin, J. Lin, and S. Han, “Haq: Hardware aware
automated quantization with mixed precision,” in Proceedings
of the IEEE Conference on Computer Vision and Pattern Recognition,
pp. 8612–8620, 2019.


\bibitem{mass}W. Maass, ``Networks of spiking neurons:  the third generation of neural network models,'' Neural networks, 10(9):1659–1671, 1997.

\bibitem{esser}S. K. Esser, P. A. Merolla, J. V. Arthur, A. S. Cassidy, R. Appuswamy,
A. Andreopoulos, D. J. Berg, J. L. McKinstry, T. Melano, D. R. Barch,
C. di Nolfo, P. Datta, A. Amir, B. Taba, M. D. Flickner, and D. S.
Modha, “Convolutional networks for fast, energy-efficient neuromorphic
computing,” Proceedings of the National Academy of Sciences, vol. 113,
number 41, pp. 11441–11446, 2016.

\bibitem{dora}S. Dora, S. Sundaram, and N. Sundararajan, “A two stage learning
algorithm for a growing-pruning spiking neural network for pattern classification
problems,” in Neural Networks (IJCNN), 2015 International
Joint Conference On, pp. 1–7, IEEE, 2015.

\bibitem{shi}Y. Shi, L. Nguyen, S. Oh, X. Liu, and D. Kuzum, “A soft-pruning
method applied during training of spiking neural networks for in memory
computing applications,” Frontiers in neuroscience, vol. 13,
p. 405, 2019.
\bibitem{rathi}N. Rathi, P. Panda, and K. Roy, “Stdp based pruning of connections
and weight quantization in spiking neural networks for energy-efficient
recognition,” IEEE Transactions on Computer-Aided Design of Integrated
Circuits and Systems, 2018.

\bibitem{yousefzadeh} A. Yousefzadeh, E. Stromatias, M. Soto, T. Serrano-Gotarredona, and
B. Linares-Barranco, “On practical issues for stochastic stdp hardware
with 1-bit synaptic weights,” Frontiers in neuroscience, vol. 12, 2018.
\bibitem{gopal} G. Srinivasan and K. Roy, “Restocnet: Residual stochastic binary convolutional
spiking neural network for memory-efficient neuromorphic
computing,” Frontiers in Neuroscience, vol. 13, p. 189, 2019.


\bibitem{admm}L. Deng, Y. Wu, Y. Hu, L. Liang, G. Li, X. Hu, Y. Ding, P. Li, and Y. Xie, ``Comprehensive snn compression using admm optimization and activity regularization,'' arXiv preprint arXiv:1911.00822, 2019.

\bibitem{spike-thrift}S. Kundu, G. Datta, M. Pedram, and P. A. Beerel,  ``Spike-thrift: Towards energy-efficient deep spiking neural networks by limiting spiking activity via attention-guided compression," In Proceedings of the IEEE/CVF Winter Conference on Applications of Computer Vision (pp. 3953-3962), 2021.

\bibitem{hybrid}N.  Rathi,  G. Srinivasan,  P.  Panda,  and  K.  Roy, ``Enabling  deepspiking neural networks with hybrid conversion and spike timing dependent backpropagation,''  In International Conference on Learning Representations, 2020.  URL https://openreview.net/forum?id=B1xSperKvH.

\bibitem{rbodo}B.  Rueckauer,  I.  Lungu,  Y.  Hu,  M. Pfeiffer,  and  S.  Liu, ``Conversion  of  continuous-valued  deep  networks  to  efficient  event-driven  networks  for  image classification,'' Frontiers in neuroscience, 11:682, 2017.

\bibitem{kmeans}J. MacQueen, ``Some methods for classification and analysis of multivariate observations." Proceedings of the fifth Berkeley symposium on mathematical statistics and probability. Vol. 1. No. 14. 1967.


\bibitem{eliasmith}E. Hunsberger  and  C. Eliasmith, ``Spiking  deep  networks  with  lif  neurons," arXiv  preprint arXiv:1510.08829, 2015.

\bibitem{diehl2015fast}P. U. Diehl, D. Neil, J. Binas, M. Cook, S. Liu, and M. Pfeiffer, ``Fast-classifying, high-accuracy spiking deep networks through weight and threshold balancing," In 2015 International Joint Conference on Neural Networks (IJCNN), pp. 1–8. IEEE, 2015.

\bibitem{sengupta2019going}A. Sengupta, Y. Ye, R. Wang, C. Liu, and K. Roy, ``Going deeper in spiking neural networks: Vgg and residual architectures," Frontiers in neuroscience, 13:95, 2019.

\bibitem{neftci2019surrogate}E. Neftci, H. Mostafa, and F. Zenke,  ``Surrogate gradient learning in spiking neural networks," IEEE Signal Processing Magazine, 36:61–63, 2019.



\bibitem{bellec2018long}G. Bellec, D. Salaj, A. Subramoney, R. Legenstein, and W. Maass, ``Longshort-term memory and learning-to-learn in networks of spiking neurons,''  In Advances in Neural Information Processing Systems, pp. 787–797, 2018.

\bibitem{pca}I. Garg, P. Panda, K. Roy, ``A low effort approach to structured cnn design using pca,'' IEEE Access, 8, 1347-1360, 2019.

\bibitem{deepcompression}S. Han, H. Mao, and W. J. Dally, ``Deep compression: Compressing deep neural networks with pruning, trained quantization and huffman coding," arXiv preprint arXiv:1510.00149, 2015.



\bibitem{leak}S. S. Chowdhury, C. Lee, and K. Roy, ``Towards Understanding the Effect of Leak in Spiking Neural Networks,'' arXiv preprint arXiv:2006.08761, 2020.

\bibitem{dctsnn}I. Garg, S. S. Chowdhury, and K. Roy,  ``DCT-SNN: Using DCT to Distribute Spatial Information over Time for Learning Low-Latency Spiking Neural Networks,'' arXiv preprint arXiv:2010.01795, 2020.

\bibitem{distill2}J. H. Cho, and B. Hariharan, ``On the efficacy of knowledge distillation,'' In Proceedings of the IEEE/CVF International Conference on Computer Vision (pp. 4794-4802), 2019.

\bibitem{horowitz20141} M. Horowitz, ``1.1 computing’s energy problem (and what we can do about it),"  In 2014 IEEE International  Solid-State  Circuits  Conference  Digest  of  Technical  Papers  (ISSCC),  pp.  10–14, IEEE, 2014.

\bibitem{nettrim}A. Aghasi, A. Abdi, N. Nguyen, and J. Romberg, ``Net-trim: Convex pruning of deep neural networks with performance guarantee,'' arXiv preprint arXiv:1611.05162, 2016.

\bibitem{saima}S. Sharmin, N. Rathi, P. Panda, and K. Roy, ``Inherent adversarial robustness of deep spiking neural networks: Effects of discrete input encoding and non-linear activations,'' In European Conference on Computer Vision, pp. 399-414, Springer, Cham, 2020.

\bibitem{ttfs}S. Park, S. Kim, B., Na, and S. Yoon, ``T2FSNN: deep spiking neural networks with time-to-first-spike coding,'' In 2020 57th ACM/IEEE Design Automation Conference (DAC), pp. 1-6, IEEE, 2020.

\bibitem{phase}J. Kim, H. Kim, S. Huh, J. Lee, and K. Choi,  ``Deep neural networks with weighted spikes,'' Neurocomputing, 311:373–386, 2018.

\bibitem{burst}S. Park, S. Kim, H. Choe, and S. Yoon,  ``Fast and efficient information transmission with burst spikes in deep spiking neural networks,''  In 2019 56th ACM/IEEE Design Automation Conference (DAC), pp. 1–6. IEEE, 2019.




\bibitem{cao}Y. Cao, Y. Chen, and D. Khosla,  ``Spiking deep convolutional neural networks for energy-efficient  object  recognition,'' International  Journal  of  Computer  Vision,  113(1):54–66, 2015.

\bibitem{wu18}Y. Wu,  L. Deng,  G. Li,  J. Zhu,  and L. Shi, ``Spatio-temporal backpropagation for training high-performance spiking neural networks,'' Frontiers in neuroscience, 12:331, 2018.

\bibitem{distill}R. K. Kushawaha, S. Kumar, B. Banerjee, and R. Velmurugan, ``Distilling Spikes: Knowledge Distillation in Spiking Neural Networks,'' arXiv preprint arXiv:2005.00288, 2020.

\bibitem{lu}S. Lu, and A. Sengupta,`` Exploring the connection between binary and spiking neural networks,'' Frontiers in Neuroscience, 14, 535, 2020.




\end{thebibliography}
\end{document}